\documentclass[sigconf,natbib=false, anonymous]{acmart}
\newcommand{\etal}{\textit{et al.}}
\usepackage{pifont}
\usepackage{cleveref}
\usepackage{multirow}
\usepackage{hyperref}

\AtBeginDocument{%
  }

\setcopyright{acmcopyright}
\copyrightyear{2023}
\acmYear{2023}

\acmConference[MM '23]{the 31st ACM International Conference on Multimedia}{October 29, 2023}{Ottawa, Canada}

\acmSubmissionID{712}
\RequirePackage[
  datamodel=acmdatamodel,
  style=acmnumeric,
  ]{biblatex}

\begin{document}

\title{Flare-Aware Cross-modal Enhancement Network for Multi-spectral Vehicle Re-identification}

\author{Aihua~Zheng}
\authornote{Both authors contributed equally to this research.}
\email{ahzheng214@foxmail.com}
\orcid{1234-5678-9012}
\author{G.K.M. Tobin}
\authornotemark[1]
\email{webmaster@marysville-ohio.com}
\affiliation{%
  \institution{Institute for Clarity in Documentation}
  \streetaddress{P.O. Box 1212}
  \city{Dublin}
  \state{Ohio}
  \country{USA}
  \postcode{43017-6221}
}

\author{Zhiqi~Ma}
\affiliation{%
  \institution{Anhui University}
  \streetaddress{1 Th{\o}rv{\"a}ld Circle}
  \city{Hefei}
  \country{Iceland}}
\email{doermzq0398@foxmail.com}

\author{Zi~Wang}
\affiliation{%
  \institution{Inria Paris-Rocquencourt}
  \city{Rocquencourt}
  \country{France}
}
\email{ziwang1121@foxmail.com}

\author{Chenglong~Li}
\affiliation{%
  \institution{Inria Paris-Rocquencourt}
  \city{Rocquencourt}
  \country{France}
}
\email{lcl1314@foxmail.com}

\renewcommand{\shortauthors}{Zheng et al.}
\begin{abstract}

Multi-spectral vehicle Re-identification (Re-ID) aims to incorporate complementary visible and infrared information to tackle the challenge of re-identifying vehicles in complex lighting conditions.
However, in harsh environments, the discriminative cues in RGB and NI (near infrared) modalities are always lost by the strong flare from vehicle lamps or the sunlight, and existing multi-modal fusion methods can not recover these important cues and thus have limited performance. 
%
%
To handle this problem, we propose a \textbf{F}lare-\textbf{A}ware \textbf{C}ross-modal \textbf{E}nhancement Network (\textbf{FACENet}) to adaptively restore the flare-corrupted RGB and NI features with guidance from the flare-immunized TI (thermal infrared) spectrum. 
First, to reduce the influence of locally degraded appearance by the intense flare,
we propose a \textbf{M}utual \textbf{F}lare \textbf{M}ask \textbf{P}rediction (\textbf{MFMP}) module to jointly obtain the flare-corrupted masks in RGB and NI modalities in a self-supervised manner.
%
Second, to utilize the flare-immunized TI information to enhance the masked RGB and NI, 
we propose a \textbf{F}lare-aware \textbf{C}ross-modal \textbf{E}nhancement module (\textbf{FCE}) to adaptively guide feature extraction of masked RGB and NI spectra with the prior flare-immunized knowledge from the TI spectrum.
%
Third, 
to mine the common informative semantic information of RGB and NI, we propose an \textbf{I}nter-modality \textbf{C}onsistency (\textbf{IC}) loss to enforce the semantic consistency between the two modalities.
Finally, to evaluate the proposed FACENet while handling the intense flare problem, we contribute a new multi-spectral vehicle Re-ID dataset, named WMVEID863 with additional challenges, such as motion blur, huge background changes, and especially intense flare degradation.
Comprehensive experiments on both the newly collected dataset and public benchmark multi-spectral vehicle Re-ID datasets verify the superior performance of the proposed FACENet compared to the state-of-the-art methods, especially in handling the strong flares. 
\textcolor{blue}{The codes and dataset will be released at \href{https://github.com/Mzq12138/Official-Implementation-for-Flare-Aware-Cross-modal-Enhancement-for-Multi-spectral-Vehicle-ReID.}{\textbf{this link}}}.
\end{abstract}
\maketitle






\section{Introduction}
\label{sec:intro}

%
Multi-spectral vehicle Re-ID endeavors to address the challenge of re-identifying vehicles (Re-ID) in intricate lighting conditions by introducing complementary near-infrared (NI) and thermal infrared (TI) data.
Li \etal~\cite{li2020multi} first propose multi-spectral vehicle Re-ID dataset RGBNT100 and RGBN300 with the baseline method HAMNet to maintain the similarity between heterogeneous spectra and automatically fuse different spectrum features in an end-to-end manner.
Zheng \etal~\cite{zheng2022multi} propose to simultaneously tackle the discrepancies from both modality and sample aspects with an adaptive layer normalization unit to handle intra-modality distributional discrepancy and contribute a high-quality dataset MSVR310.

\begin{figure}[H]
\centering
\begin{center}
  \includegraphics[width=1\linewidth]{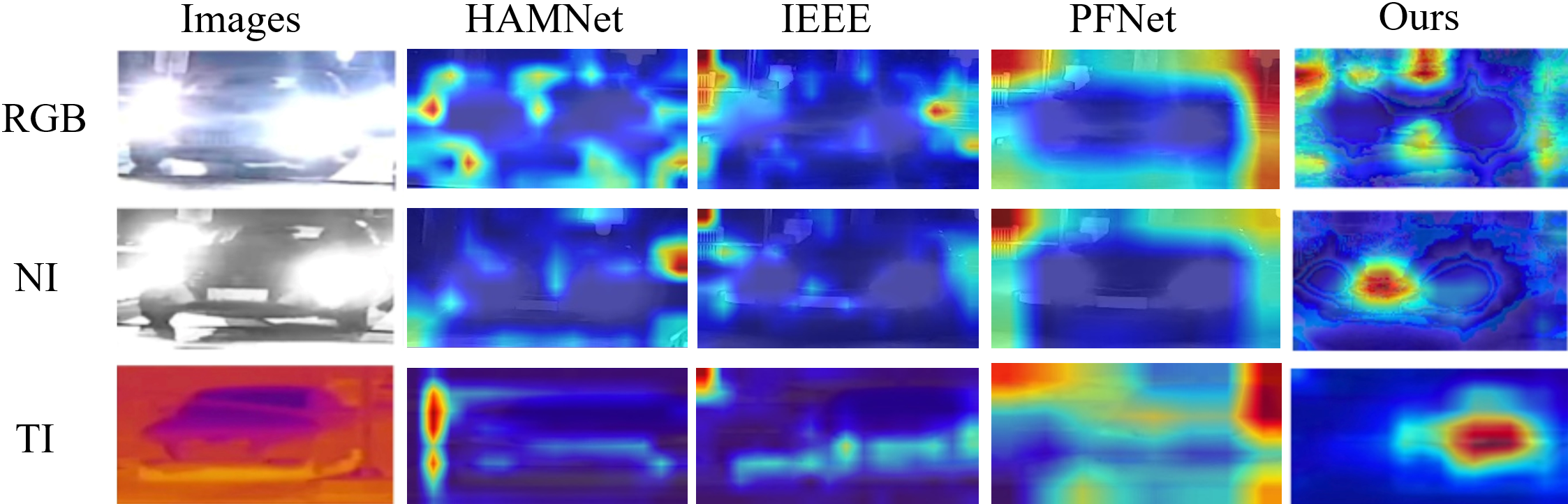}
  \end{center}
  \caption{
  Feature map visualization of how the intense flares affect existing multi-modal methods.
  }
  \label{fig:motivation}
\end{figure}

Despite the progress in both methodology and dataset, the ubiquitous intense flare problem is neglected in real-life complex transportation systems, 
since both RGB and Near-Infrared (NI) modalities are sensitive to intense flares such as vehicle lamps or strong sunlight.
As a result, discriminative cues in RGB and NI images are partially lost, which significantly affects the modality fusion in existing multi-modal methods.
The most straightforward solution is to employ image restoration strategies~\cite{ZamirAKHKYS20, singh2020single} to restore the local region of the light-corrupted images.
However, modeling a projection between light-corrupted and clean images requires a large amount of paired data, which is tedious at best and impossible at worst.
Multi-spectral Re-identification (Re-ID) methods are capable of effectively overcoming illumination limitations in a specific spectrum by fusing or enhancing complementary spectra.
However, most of the existing multi-spectral Re-ID methods explore the global information enhancement among complementary multi-modalities to address low-illumination issues.
While re-identifying the vehicles with intense flare, there emerges additional local severe information degradation, which can make existing methods unreliable.
Therefore, how to effectively utilize the complementary multi-spectral information to guide the local enhancement with the intense flare while simultaneously maintaining the global feature fusion without intense flare remains unstudied for multi-spectral vehicle Re-identification.

As a new challenge task in multi-spectral vehicle Re-ID, we observe that there are three crucial issues concerning the intense flare problem.
\textbf{First}, to eliminate the impact of the local information degradation from the intense flare, it is straightforward to locate the flare-corrupted region in the images.
Meanwhile, different modalities, such as RGB and NI, may produce discrepant degradation under the intense flare.
Therefore, how to automatically and jointly predict the flare-corrupted local region in both RGB and NI modalities is essential for this task. 
\textbf{Second}, we observe that the thermal-infrared (TI) spectrum is generally flare-immunized, which can provide critical information, especially for the flare-corrupted region in RGB and NI modalities.
Therefore, how to utilize the flare-immunized TI information to guide the feature learning of the masked RGB and near-infrared (NI) spectra is crucial.
\textbf{Third}, after the cross-modal enhancement, the guidance of the TI spectrum may introduce unexpected modality-specific information to RGB and NI modalities, leading to a biased feature distribution.
Therefore, intuitively enforcing the three modalities such as 3M loss~\cite{wang2022interact} and CdC loss~\cite{zheng2022multi}  may result in suboptimal performance.
%
%
How to align the semantic information of enhanced RGB and NI features is essential for multi-spectral vehicle Re-ID.
%
%

To solve the above problems, we propose Flare-Aware Cross-modal Enhancement Net (FACENet) to adaptively restore the flare-corrupted RGB and NI features with the guidance of the flare-immunized TI spectrum.
\textbf{First}, to reduce the influence of locally degraded appearance by the intense flare, we propose a Mutual Flare Mask Prediction (MFMP) module to jointly obtain the  flare-degraded regions in both RGB and NI in a self-supervised manner.
MFMP contains two main components, 1) a Self-supervised Mask Prediction (SMP) scheme that supervises the flare mask with the pseudo-label according to the image histogram,
and 2) a Flare-susceptible Modality Interaction (FMI) scheme to mutually interact the common flare-affected features in RGB and NI for more robust mask prediction.
\textbf{Second}, although we can relieve the influence of the intense flare from the masked images, the crucial local information is simultaneously missing with the masks.
Therefore, we propose a Flare-aware Cross-modal Enhancement (FCE) module to adaptively guide the feature extraction of the masked RGB and NI spectra with the prior
flare-immunized knowledge from the TI spectrum in a conditional learning manner.
\textbf{Third}, 
%
%
%
To utilize the common representative information in RGB and NI, and simultaneously avoid the over-intervention of the TI spectrum, we propose an inter-modality consistency loss (IC Loss) to enhance the semantic feature extraction and constrain the consistency between RGB and NI spectra to suppress the over-effect of the TI spectrum.
Note that, when there is no intense flare, we can switch off the FCE module as shown in Fig.~\ref{fig:network}.
In this way, the features of RGB and NI will be replaced by the output of Flare-susceptible Modality Interaction (FMI).
%
%

In addition, existing multi-spectral vehicle Re-ID datasets, including MSVR310~\cite{zheng2022multi}, RGBN300, and RGBNT100 \cite{li2020multi} mainly focus on low illumination scenarios with still vehicles and without intense flares.
Furthermore, the limited amount of identities and image samples tend to overfit.
%
%
%
%
%
%
Therefore, we contribute a more realistic large-scale {\textbf{W}}ild {\textbf{M}}ulti-spectral {\textbf{Ve}}hicle Re-{\textbf{ID}} dataset WMVeID863 in a complex environment in this paper.
WMVeID863 is captured with vehicles in motion with more challenges, such as motion blur, huge background changes, and especially intense flare degradation from car lamps, and sunlight. 
It contains 863 identities of vehicle triplets (RGB, NI, and TI) captured with 8 camera views at a traffic checkpoint, contributing 14127 images with the vehicles in motion.
More detailed information on WMVeID863 and the comparison with existing datasets can be referred to in Section 4.

%
%
%
We conclude our main contributions as follows. 

%
%
%
%
\begin{itemize}
    \item We are the first to launch the intense flare issue in vehicle Re-ID and  propose the Flare-Aware Cross-modal Enhancement Network (FACENet) to adaptively restore the flare-corrupted RGB and NI features with guidance from the flare-immunized TI spectrum.
    \item To reduce the influence of locally degraded appearance by the intense flare, we propose the Mutual Flare Mask Prediction (MFMP) module to predict the flare mask in a self-supervised manner (SMP) and interact the common flare-affected features in RGB and NI modalities (FMI).
    \item To effectively utilize the flare-immunized TI information to enhance the masked RGB and NI features, We propose the Flare-aware Cross-model enhancement (FCE) module to guide local feature learning of RGB and NI branches with prior information from the flare-immune branch. 
    \item To explore the semantic consistency in multi-spectral vehicle Re-ID with intense flare, we propose the Inter-Modality Consistency (IC) loss between the enhanced RGB and NI spectra by minimizing the KL divergence.
    \item We contribute a realistic large-scale Wild Multi-spectral Vehicle Re-ID dataset WMVeID863 with more challenges including intense flare to evaluate the effectiveness of FACENet. Comprehensive experiments verify the superior performance of the proposed FACENet against the state-of-the-art methods, especially while handling the intense flare problem.
    
\end{itemize}

\section{Related Work}

\subsection{Multi-spectral Re-ID}
Despite the recent progress on visible vehicle Re-ID~\cite{liu2016deep,liu2016deepID,li2020multi,meng2020parsing,chen2020orientation,zhao2021phd,chen2022sjdl,li2022attribute},
multi-spectral vehicle Re-ID emerges by utilizing the complementary visible (RGB), near-infrared (NIR), and thermal-infrared (TIR) spectra information to handle the harsh and complex lighting conditions for vehicle Re-ID.
%
%
%
%
%
Li \etal~\cite{li2020multi} construct the first multi-spectra vehicle Re-ID dataset RGBN300 (visible and near-infrared) and RGBNT100 (visible, near-infrared, and thermal infrared) to solve this problem and proposed a baseline method HAMNet to effectively fuse multi-spectra information through CAM (Class Activation Map)~\cite{zhou2016learning}. 
Zheng \etal~\cite{zheng2022multi} propose a Cross-directional Consistency Network to solve the huge cross-modal discrepancy caused by different views and modalities and contribute a high-quality multi-spectral vehicle Re-ID benchmark MSVR310.
Meanwhile,
Zheng \etal~\cite{zheng2021robust} construct a multi-spectral person Re-ID dataset RGBNT201 and a progressive fusion network for robust feature extraction. 
Wang \etal~\cite{wang2022interact} further consider the  spectra-specific information and propose to boost specific information via incorporating other spectra in the fusing phase for multi-spectral person Re-ID.
However, existing multi-spectral Re-ID methods primarily focus on utilizing global complementary multi-modality information. However, they remain unreliable in addressing the issue of intense flares and local information degradation.
Therefore, it is ineffective to directly employ the existing multi-spectral Re-ID methods for the intense flare issue.

\subsection{Cross-modal Enhancement}
Cross-modal enhancement is committed to boosting one modality from other modalities.
Wang \etal~\cite{crossenhance} propose to enhance text representations by integrating visual and acoustic information into a language model.
Wang \etal~\cite{wang2022interact} propose to exchange the information between modalities to absorb complementary information from other modalities while simultaneously maintaining the modality-specific information.
Cheng \etal~\cite{cheng2020look} propose a self-supervised framework with a co-attention mechanism to learn generic cross-modal representations to solve the pretext task of audiovisual synchronization.
Mercea \etal~\cite{mercea2022audio} propose to learn multi-modal representations from audiovisual data using cross-modal attention for the alignment between the audio and visual modalities.
%
Zheng \etal~\cite{zheng2021robust} integrate the multi-modal features at part-level to capture the complementary local information among modalities.
However, existing cross-modal enhancement methods are limited while facing flare problems on certain spectra, since the intense flare introduces large influence and diversity in different modalities.
Therefore, directly performing cross-modal enhancement will bring noise and reduce the representation ability of RGB and NI spectra due to the heterogeneity of multi-spectra.

\section{Proposed Method}
\begin{figure*}[!ht]
\centering
\begin{center}
  \includegraphics[width=\linewidth]{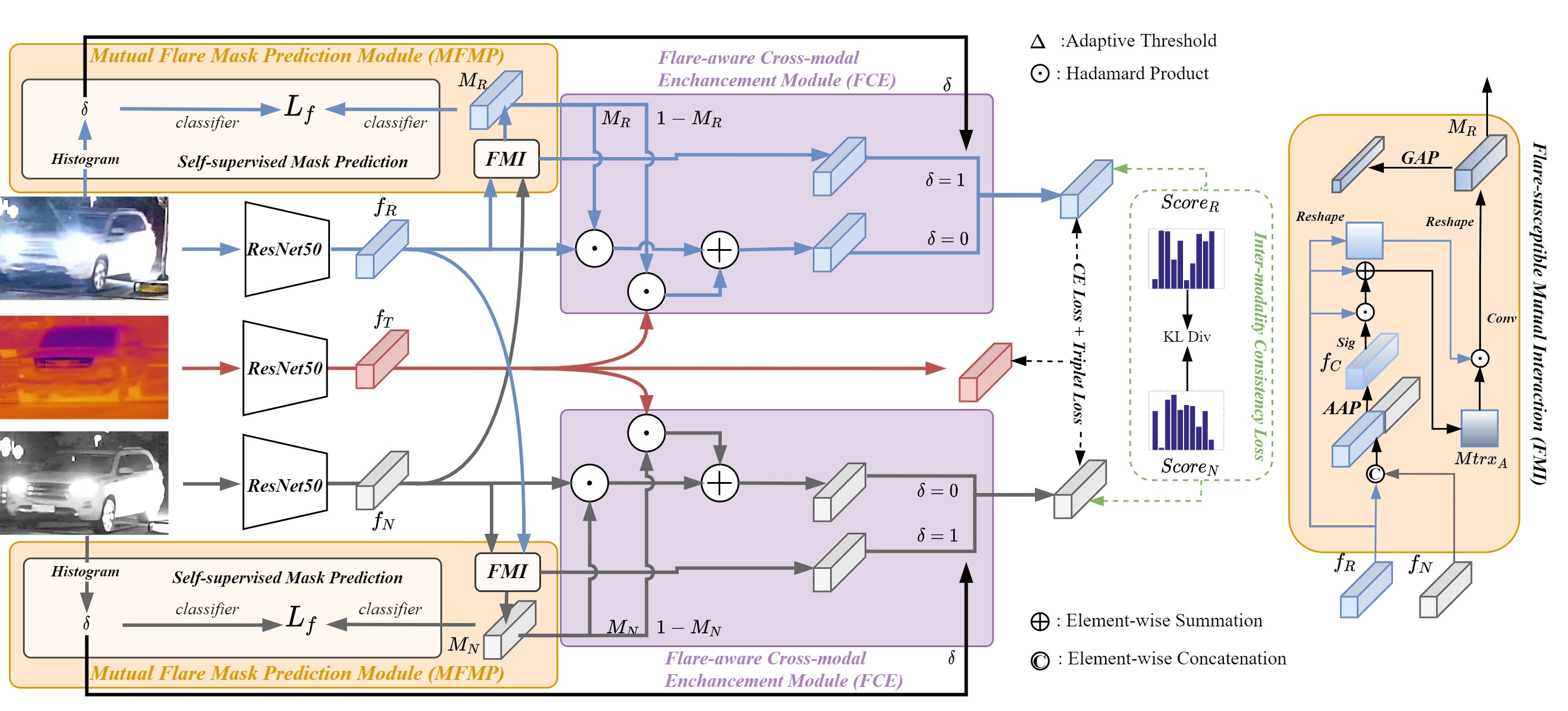}
  \end{center}
  \caption{
  The overall structure of the proposed method FACENet. 
  First, we obtain the pseudo-label according to the histogram.
  We extract features of multi-spectrum vehicle images by three individual ResNet50~\cite{he2016deep} to obtain features $f_{R}$, $f_{N}$, $f_{T}$ of each spectrum.
  $f_{R}$ and $f_{N}$ are fed into the FMI to obtain the flare masks for each spectrum.
  Then, we train the mutual flare mask prediction from FMI under the supervision of the pseudo-label.
  Finally, the flare-immunized $f_{T}$ is fed into a flare-aware cross-modal enhancement module to guide $f_{R}$ and $f_{N}$ with the aid of the predicted flare masks.
  The overall training stage for Re-ID is supervised by Cross-Entropy loss, and Triplet loss~\cite{triplet} on three individual spectra, together with the proposed IC loss on RGB and NI spectra. 
  In the testing stage, we concatenate the multi-branch features for final distance measuring.
        }
  \label{fig:network}
\end{figure*}
%
%
%
To make full use of the flare-immunized information in the TI spectrum and the residual effective information in the flare-corrupted RGB and NI spectra, we propose the \textbf{F}lare-\textbf{A}ware \textbf{C}ross-modal \textbf{E}nhancement \textbf{N}etwork (FACENet) for multi-spectral vehicle Re-ID, as shown in Fig.~\ref{fig:network}.
%
%
First of all, we propose the \textbf{M}utual \textbf{F}lare-aware \textbf{M}ask \textbf{P}rediction (MFMP) module to extract the regions heavily affected by the flare by \textbf{F}lare-susceptible \textbf{M}odality \textbf{I}nteraction (FMI) scheme, and the \textbf{S}elf-supervised \textbf{M}ask \textbf{P}rediction (SMP) scheme.
Then, 
we propose the \textbf{F}lare-aware \textbf{C}ross-modal \textbf{E}nhancement (FCE) module to use the local feature in TI spectra as a condition to guide the local feature recovery in RGB and NI spectra.
Moreover, to minimize the similarity between predictions of RGB and NI branch, we design \textbf{I}nter-Modality \textbf{C}onsistency Loss (IC Loss) for training.
%
We will elaborate on the main components below.

\subsection{Feature Extraction}
    We follow multi-stream network (MSN) architecture \cite{li2020multi} and adopt ResNet50 \cite{he2016deep} pre-trained on ImageNet \cite{deng2009imagenet} as the basic feature extractor for three spectra.
    It is worth noting that the three feature extractors do not share parameters.
    After obtaining the features, we first apply feature interaction through the FMI module
    The MFMP and FCE modules are plugged into the last layer of ResNet50. 
    At the training stage, we apply the CE loss to train the classifier, the Triplet loss \cite{triplet} to optimize the margin of positive and negative samples, and the proposed IC loss to align the semantic information of the enhanced RGB and NI spectra.
    At the testing stage, we concatenate the features from three branches for final Re-ID matching.

\subsection{MFMP: Mutual Flare Mask Prediction Module}

In order to reduce the influence of locally degraded appearance by the intense flare, we propose the Mutual Flare Mask Prediction (MFMP) Module to
jointly obtain the flare-corrupted masks in RGB and NI modalities
in a self-supervised manner.
MFMP module consists of a Self-supervised Mask Prediction (SMP) scheme module and a Flare-susceptible Mutual Interaction (FMI) scheme.
The SMP scheme intends to predict the region affected by flares through a self-supervised approach, whereby the pseudo label is derived from the histogram of a specific image.
However, training the mask separately can not effectively leverage the complementary information present in flare-susceptible spectra. 
Therefore, we propose the FMI scheme to integrate features from both RGB and NI to produce a more effective flare-aware mask.
%
%

\noindent
\textbf{Self-supervised Mask Prediction (SMP).}
we propose a mask prediction module in a self-supervised manner to obtain the feature mask for further cross-modal enhancement. 
Self-supervised Mask Prediction scheme (SMP) aims to extract the flare-corrupted region in a certain image of flare-susceptible spectra using the pseudo-label of a certain image. 

To supervise the optimization of the mutual flare mask prediction module, we derive a pseudo-label that indicates whether a particular image is affected by flares based on the histogram of the image. Specifically, we calculate the percentage of pixels in the image with pixel values between 250 and 255 relative to the total number of pixels in the image, which is labeled as $\delta$.
To obtain the pseudo-label, we manually select a bar $10\%$. If the percentage is greater than the bar, the image is considered a flare-corrupted sample.
The pseudo-label $\delta$ is also utilized in the FCE module to exclude samples that are not significantly affected by flares.
To supervise the FMP module, we define Binary Cross-Entropy loss $L_{bce}$~ \cite{ruby2020binary} as $L_{f}$ to perform binary classification and determine whether an image is affected by flares, defined as:
\begin{equation}
  L_{f} = - {\textstyle \sum_{j=1}^{2}} y^{n}\log_{}{\hat{y}_{j}^{n}}.
  \label{eq:bceloss}
\end{equation}

To obtain the mask, we set an adaptive threshold $\Delta$ to determine whether the flare affects a certain pixel.
The threshold is initialized to zero and updates adaptively with the optimization of the whole network.

\noindent
\textbf{Flare-susceptible Mutual Interaction (FMI).}
While a flare mask can be derived from a feature map of a single spectrum, this approach overlooks the fact that flares affect both RGB and NI spectra. 
Therefore, training the flare mask separately is suboptimal.
To address this limitation, we propose integrating features from both RGB and NI to accurately identify the local areas affected by flares.
As shown in Fig.~\ref{fig:network}, given features from RGB and NI branches $f_{R}^{B \times C \times H \times W}$ and $f_{N}^{B \times C \times H \times W}$ respectively,
we first concatenate RGB and NI in the channel dimension for the fused feature $f_{C}^{B \times 2*C \times H \times W}$, then we apply 4 convolution layers with the kernel size of $3\times3, 3\times3, 1\times1, 3\times3$, respectively to expand the receptive field while maintaining non-linear for a broad common region of flare to obtain the common feature of RGB and NI $f_{C}^{\prime B \times C \times H \times W}$. 
Then we apply adaptive average pooling (AAP) and 2 convolution layers with the kernel size of $1\times1, 1\times1$ to transform the feature to target dimension while maintaining non-linearity, plus a $Sigmoid$ function to obtain attention mask of common feature $f_{att}^{B \times C \times H \times W}$ for both RGB and NI spectrum. 

After the residual connection between the common feature $f_{C}^{\prime}$ and the affine matrix $Mtrx_{A}$ we obtain the common flare-corrupted features $f_{C}^{B \times C \times H \times W}$ of RGB and NI as shown in Equ.~\ref{eq:F_com}, where $conv_1$ and $conv_3$ indicates a convolution layer with the kernel size of $1\times1$ and $3\times3$ respectively, and $\odot$ and $\oplus$ denotes the Hadamard product and element-wise sum.
\begin{equation}
  f_{C} = concat (f_{R}, f_{N}),
  \label{eq:F_cat}
\end{equation}
\begin{equation}
  f_{C}^{\prime} = conv_{3} (conv_{3} (conv_{1} (conv_{3} (f_{C})))),
  \label{eq:F_cat'}
\end{equation}
\begin{equation}
  f_{C} = conv_{1} (conv_{1} (AAP (f_{C}^{\prime}))) \odot f_{att} \oplus f_{C}^{\prime}.
  \label{eq:F_com}
\end{equation}
%
%
%

In order to obtain the specific cross-modal attention mask of the flare-corrupted area in a certain spectrum, such as RGB spectrum, we apply convolution operation on the common features $f_{C}^{B \times C \times H \times W}$ and features of RGB $f_{R}^{B \times C \times H \times W}$ with the kernel size of $1\times1$ respectively to obtain the flatten feature matrices for spatial feature extraction.
Then we obtain an affine matrix $Mtrx_{A}^{B \times C \times H*W}$, which describes the similar relationship between complementary features of the common features and features from a certain spectrum, such as RGB,  at different spatial locations.
To further utilize the relational similarity, we obtain the $Mtrx_{A}^{B \times C \times H*W}$ through the residual connection between $f_{R}$ and $f_{C}$.
\begin{equation}
  Mtrx_{A} = M_{com} \odot f_{R} \oplus f_R.
  \label{eq:M_affine}
\end{equation}

To maintain the relative position of the channel-level flare mask $M_{R}$, the affine matrix is then reshaped to the shape of $B \times C \times H \times W$ for the residual connection with the feature matrix $ M_{R}$.
The final flare mask prediction is the residual connection between $f_{N}$ and $M_{N}^\prime$.
\begin{equation}
  M_{R} = f_{R} \odot M_{R} \oplus f_{R}.
  \label{eq:output}
\end{equation}
%


\subsection{FCE: Flare-aware Cross-modal Enhancement Module}
Once the flare mask is obtained after the MFMP module, a crucial issue for multi-spectra vehicle Re-ID under intense flare conditions is how to recover the flare-corrupted features in RGB and NI spectra.
Inspired by the conditional learning \cite{li2020segmenting, wang2018recovering}, 
which incorporates additional information such as labels, context, or prior knowledge, into the learning process to help the model better understand the input data and the specific task requirement,
we propose the Flare-aware Cross-modal Enhancement (FCE) module to use the flare-immunized TI spectrum to guide the feature extraction of the flare-corrupted RGB and NI spectra.
%

The FCE module aims to guide the flare-susceptible RGB and NI spectra with the flare-immunized information in the TI spectrum.
Therefore, we consider the local features masked by flare-mask prediction from MFMP in the TI spectrum as the condition.
And the masked features in RGB and NI as the input of conditional learning.
Note that FCE is performed respectively on RGB and NI spectra.
Taking RGB as an example of FCE:
\begin{equation}
  FCE(f_{R}) = {f_{T}}\odot{M_{R}} \oplus ({f_{R}} \odot \sim{M_{R}}).
  \label{eq:fce}
\end{equation}
where $\sim{M_{R}}$ is the negation of $M_{R}$.


Fig.~\ref{fig:network} illustrates the objective of FCE, which aims to guide the learning of flare-corrupted $f_{{S}}$ from flare-immunized features $f_{T}$, where $f_{{S}}, S \in {\{RGB, NI\}}$ indicates features from $RGB$ or $NI$ spectrum.
Specifically, the TI branch provides flare-immunized knowledge in the form of $f_{T}$, and FCE seeks to restore flare-corrupted information in $f_{S}$ by utilizing the prior knowledge.

After obtaining $M_{S}$ from MFMP, where $M_{S}, S \in {\{RGB, NI\}}$ indicates the mask value of each pixel indicates how intense a certain pixel is affected by the flare in $RGB$ or $NI$ spectrum.
Then we perform element-wise production $\odot$ on feature $f_{S} \odot M_{S}$ to obtain the local features that are affected by flare while performing $f_{T} \odot M_{S}$ to obtain the corresponding flare-immunized local feature in the TI spectrum.

At last, the two local features are summed together as the final feature representation for conditional learning of the ${S}, S \in {\{RGB, NI\}}$ branch.
\textbf{Note that} the FCE module is designed for the flare problem, for certain samples without severe flare degradation, we utilize pseudo-label to skip from FCE.

\subsection{IC Loss: Inter-modality Consistency Loss}
To make full use of the complementary information of RGB and NI spectra, and avoid the disturbance caused by the modality-specific information from the TI spectrum, we propose to align the semantic consistency of RGB and NI.
%
%
Inspired by Wang \etal~\cite{wang2022interact}, 
we propose to enforce the semantic consistency of prediction scores between RGB and NI spectra.
Specifically, we propose to use KL-divergence~\cite{kldiv} to enforce the feature distribution between two specific modalities. 
This aims to limit the over-influence of the TI modality on RGB and NI, thereby minimizing the similarity between the classification results of RGB and NI.
Phuong \etal~\cite{Phuong_2019_ICCV} verify that implying consistency loss on the later prediction layer of deep networks is better.
Given classification scores of three modalities $\{S_{RGB}, S_{NI}\}$ from corresponding branches, we employ KL-divergence~\cite{kldiv} to compute the distance.
%
%

We obtain the smaller value from the two ${KL}$ distances between RGB and NI, as the proposed inter-modality consistency loss: 

%
\begin{equation}
    \mathcal{L}_{IC} = min(\sum_{i} RGB(i) \log\left(\frac{RGB(i)}{NI(i)}\right),\sum_{i} NI(i) \log\left(\frac{NI(i)}{RGB(i)}\right)).
     \label{eq:mpcloss}
\end{equation}

By enforcing the IC loss, the network can further mine the common informative semantic information of RGB and NI, and simultaneously avoid the over-intervention of the TI spectrum for a robust multi-modal representation.


\subsection{Overall Loss Function}
In the training stage, our method is under the joint supervision of cross-entropy (CE) loss, Triplet Loss~\cite{triplet}, and the proposed IC loss.
The identity loss corresponding to the image $I_{}^{j}$ is calculated as the cross entropy between the predicting probability and the identity label $y$. 
\begin{equation}
  \mathcal{L}_{id_s}^{j} = - {\textstyle \sum_{j=1}^{C}} y^{n}\log_{}{\hat{y}_{j}^{n}},
  \label{eq:celoss}
\end{equation}
where $s \in (RGB, NI, TI)$ denotes the $s$-th branch of spectrum.
$y^{n}$ is a one-hot matrix indicates the identity label of the $n$-th sample, where $y^{n}_{i} = 0, i\in\{0,1,...,C\}$ except $y^{n}_{c} = 1$.
$\hat{y}_{j}^{n}$ is the prediction results, indicating the probability that $\hat{y}_{j}^{n}$ is of the j-th class.
The identity loss for the whole network is defined as:
\begin{equation}
  \mathcal{L}_{id} = \sum^{S}_{s}\sum^{N}_{j=n} L_{id_s}^{j},
  \label{eq:celossall}
\end{equation}
where $N$ is the total number of images.

To perform hard sample mining, we randomly select $N$ identities with $P$ images in a batch and adopt triplet loss\cite{triplet} defined as:
\begin{equation}
\mathcal{L}_{tri_s} = \sum_{i=1}^{N} \sum_{j=1}^{P} max(m + max D(f_a^i, f_p^i) - minD(f_a^i, f_n^i) ,0).
\label{eq:trplt}
\end{equation}
where $m$ denotes the margin, $f$ indicates output features, $a, p, n$ denotes anchor, positive, and negative respectively, $D(f_a, f_p)$ indicates the feature distance between $f_a$ and $f_b$.
The triplet loss for the whole network is defined as:
\begin{equation}
  \mathcal{L}_{tri} = \sum^{S}_{s}\sum^{N}_{j=n} L_{tri_s}.
  \label{eq:trpltall}
\end{equation}

We train the multiple branches with binary CE loss and triplet loss~\cite{triplet}, the overall loss is defined as:
\begin{equation}
  \mathcal{L}_{all} = \mathcal{L}_{tri} + \mathcal{L}_{id} + \mathcal{L}_{f} + \mathcal{L}_{IC}.
  \label{eq:netall}
\end{equation}
%

The CE loss in FACENet is effective in distinguishing between identities, while the Triplet loss optimizes the inner-class and intra-class distances. 
In addition, our proposed IC loss adjusts the modality distance, resulting in a more robust feature for Re-ID.
\section{WMVeID863: Wild Multi-spectral Vehicle Re-identification Dataset}

To evaluate the proposed method while handling the intense flare issue, we contribute a Wild Multi-spectral Vehicle re-IDentification Dataset, named WMVeID863.
%

\subsection{Data Acquisition}
WMVeID863 dataset is collected on campus by triplicated cameras to simultaneously record RGB, NI, and TI video data in both day and night with a two-month time span.
The dataset is captured in four different weather conditions including sunny, cloudy, windy, and hot days in both morning and night respectively in videos.
%
%
The raw data contributes to 37 hours of videos in total, with RGB, NI, and TI, respectively.
The RGB and NI images are captured by the paired 360 D866 cameras with a resolution of 1920$\times$1080 in 15 fps, and the TI images are captured by a DALI thermal telescope and a HIKVISION DS-7800HQH-K1 recording device with a resolution of 1280$\times$720 in 15 fps.
To unify the resolution, we manually resize and align the RGB and NI images according to the resolution of TI images.
Then we select the video clips of vehicles with a near-front viewpoint and annotate the label according to the license numbers.
Then we crop each sample to obtain the bounding boxes of the vehicles for Re-ID.
\subsection{Data Description}
\begin{figure}[t]
\centering
\begin{center}
  \includegraphics[width=0.8\linewidth]{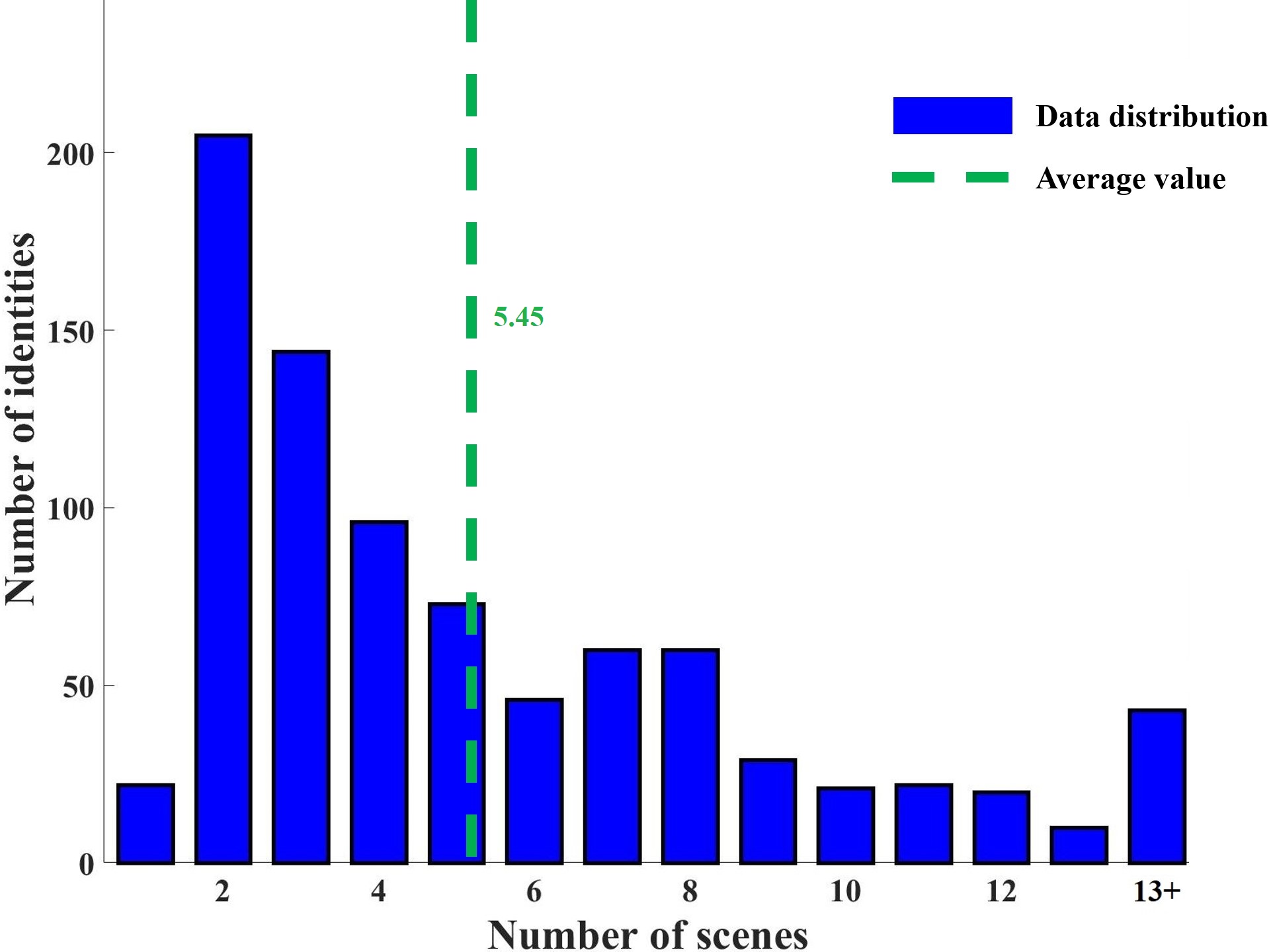}
  \end{center}
  \caption{Distribution for the number of scenes across the number of identities in WMVeID863.}
  \label{fig:data_stat}
\end{figure}

WMVeID863 contains 4709 image triplets of 863 IDs of 8 camera views, the number of image triplets of each vehicle varies from 1 to 39.
The dataset distribution of the number of scenes across the number of identities is shown in Fig.~\ref{fig:data_stat}.
We randomly select 575 IDs with 3314 image triplets for training and 288 IDs with 1309 image triplets for testing. 
The gallery samples consist of all 288 IDs with 1309 image triplets.
The query samples are randomly selected from the 288 gallery samples, consisting of 205 IDs with 928 image triplets.

\begin{table}[ht]
\centering
\setlength{\tabcolsep}{0.5mm}
\caption{Comparison with existing multi-spectra vehicle Re-ID datasets.} 
{
%

\begin{tabular}{c|c|c|c|c|c|c|c|c|c}
\hline
Benchmarks      & IDs    & Images      &Scenes  & ViM &VC & PO & VR &IL &MB   \\ \hline
RGBN300~\cite{li2020multi}         & 300   & 100250          &2004 & -&\ding{51}&\ding{51}&\ding{51}&-&-\\
RGBNT100~\cite{li2020multi}     & 100   & 17250 &690  & -  &\ding{51} &\ding{51} & \ding{51}& - &- \\
MSVR310~\cite{zheng2022multi}        & 310   & 6261        &2061 & -  &\ding{51} &\ding{51} &\ding{51}&- &-    \\
WMVeID863        & 863   & 14127       &4709 & \ding{51}  &\ding{51} &\ding{51} &\ding{51}& \ding{51} &\ding{51}   \\ \hline
\end{tabular}
}
\label{tab:dataset_comp}
\end{table}

\begin{figure}[!ht]
\centering
\begin{center}
  \includegraphics[width=1\linewidth]{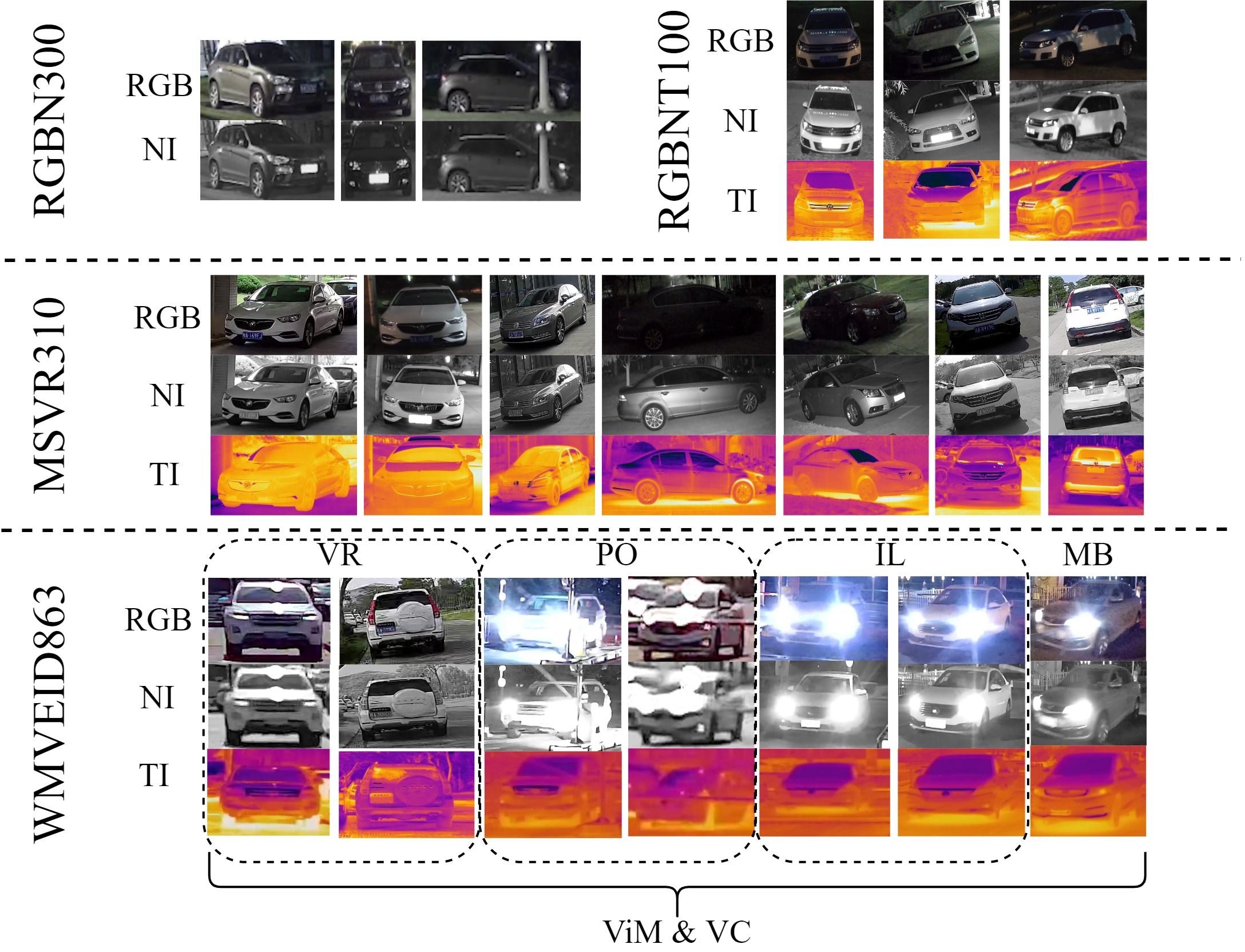}
  \end{center}
   \caption{
   Challenge comparison between the proposed WMVeID863 and existing multi-spectral datasets. Each column represents an image triplet of a vehicle ID while the two columns with the same challenges such as VR, PO, and IL derive from two different scenes of the same vehicle.
   WMVeID863 presents a more challenging and realistic multi-spectral vehicle Re-ID scenario.
   }
  \label{fig:data_challenge}
\end{figure}

 We acquire the data in a more complex environment with intense illumination from vehicle lamps or strong sunlight.
Therefore, as shown in Fig.~\ref{fig:data_challenge}, in addition to the common challenges like view changes (VC), partly occlusion (PO), and various resolutions (VR),
WMVeID863 introduces new challenges like intense light (IL).
Furthermore, different from the existing datasets that capture the vehicles in still, we capture the vehicles in motion (ViM), thereby it brings the additional challenge of motion blurring (MB).
Comparing with existing multi-spectra Re-ID datasets as reported in Table~\ref{tab:dataset_comp}, WMVeID863 has the following major advantages:
\begin{itemize}

    \item It contains a more reasonable amount of vehicle images aligned in three spectra captured with eight non-overlapping camera views. 
    WMVeID863 is the largest multi-spectral vehicle Re-ID dataset so far with more realistic challenges compared to MSVR310 ~\cite{zheng2022multi}, RGBN300~\cite{li2020multi} and RNBNT100\cite{li2020multi}.
    \item It includes image triplets of vehicles in motion captured at a traffic checkpoint, bringing more background changes and realistic challenges like intense light and motion blur, which are significantly ignored in existing multi-spectral vehicle Re-ID datasets.
\end{itemize}

\section{Experiments}
\noindent
\textbf{Implementation Details.}
We employ ResNet50~\cite{he2016deep} pretrained on ImageNet~\cite{deng2009imagenet} as our backbone. The implementation platform is PyTorch 1.0.1 with one NVIDIA RTX 3090 GPU.
We use Adam~\cite{adam} optimizer to optimize the network with the initial learning rate as $3.5\times10^{-4}$ which decays to $3.5\times10^{-5}$ and $3.5\times10^{-6}$ at 40-th epoch and 70-th epoch respectively with total 120 epochs.
The images are resized to 128$\times$256 as the input of the network.
In the training phase, the features of multiple spectra are trained separately without parameter sharing and jointly supervised by Cross-Entropy loss and Triplet Loss, while RGB and NI spectra are trained by an additional proposed IC loss.
In evaluation, we concatenate the features extracted from three parallel branches as the final representation for a sample, for the feature matching.

\noindent
\textbf{Evaluation Protocols}
Following the protocols in \cite{market1501}, we employ the commonly used metrics in Re-ID task mean Average Precision (mAP) and Matching Characteristic curve (CMC) to evaluate our method, where R-$n$ indicates the first $n$ closest samples to the query sample with the same ID from different cameras according to Euclidean distance.
\subsection{Ablation Study}
\begin{figure}[H]
\centering
\begin{center}
  \includegraphics[width=1.0\linewidth]{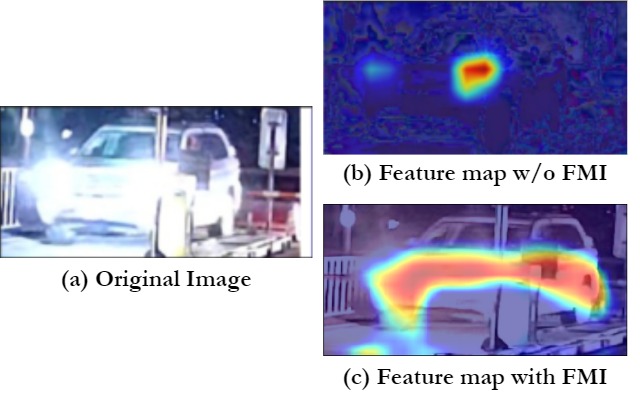}
  \end{center}
  \caption{Visualization of feature map of SMP.
  Mask prediction without FMI concentrates on the small region near the car lamp.
  With the aid of FMI, the mask prediction focuses on a much broader flare-corrupted area.}
  \label{fig:flarevisualization}
\end{figure}

We first evaluate the contribution of the three main components, mutual flare mask prediction module (MFMP), flare-aware cross-modal enhancement module (FCE), and inter-modality consistency loss (IC loss), as shown in Table~\ref{tab:ablation_study}.
First, simply introducing MFMP without flare-susceptible Mutual Interaction (FMI) into the baseline  (ResNet50 with CE loss and triplet loss) brings no significant improvement, the main reason is the flare mask simply predicted by individual RGB and NI features only contains a small part of the flare-corrupted region.
By further introducing the proposed flare-aware cross-modal enhancement module (FCE), both mAP and the R-1 score Substantially upgrade, which evidences the effectiveness of FCE to upgrade the ability of local feature guidance to the RGB and NI branches.
Removing the Flare-susceptible Modality Interaction (FMI) scheme in the MFMP module, in other words, directly performing FCE and MFMP without the Flare-susceptible Modality Interaction (FMI) scheme hinder the performance.
Together with the flare mask prediction results shown in Fig.~\ref{fig:flarevisualization} we can see that predicting the flare masks with a single spectrum without modality interaction produces biased prediction, which evidences the necessity of the proposed FMI.
Finally, enforcing IC loss $\mathcal{L}_{IC}$ further boosts the performance by considering the semantic consistency in the enhanced RGB and NI modalities.
%
%
%

%
\begin{table}[!ht]

\caption{Ablation study of FACENet on WMVeID863 (in \%).}

\centering
\begin{tabular}{l|c|c|c|c}
\hline
        Method            & mAP                         & R-1                        & R-5                      & R-10                       \\ \hline
Baseline           & 53.7                         & 60.3                         & 69.1                       & 74.4                         \\
+ MFMP (SMP + FMI)          & 53.9                         & 59.1                         & 72.2                       & 77.6                         \\
+ MFMP + FCE   & 60.5                        &  66.5                      & 76.2                         & 79.7\\ 
+ MFMP (w/o FMI) + FCE         & 57.5                         & 64.9                        & 73.3                        & 76.3 \\ 
+ MFMP + FCE + $\mathcal{L}_{IC}$  & 62.1                         &  67.3                       & 76.7                         & 80.1\\ 
\hline
\end{tabular}
\label{tab:ablation_study}
\end{table}

\subsection{Evaluation on MFMP and FCE}
\noindent
\textbf{Comparison with de-flare methods.}
To demonstrate the effectiveness of our proposed  MFMP and FCE module while handling the intense flare problem, we compare our method to two de-flare methods, contrast limited adaptive histogram equalization (CLAHE)~\cite{Zuiderveld94}, and back-projected pyramid network (BPPNet)~\cite{singh2020single}.
Note that CLAHE~\cite{Zuiderveld94} is a digital imaging processing method to enhance the image quality, and BBPNet~\cite{singh2020single} is an image restoration method to synthesize the de-flared images.
%
%
We use the images produced by the model to replace the original images for training and testing. 
The experimental results of the comparison with other de-flare methods are shown in Table~\ref{tab:comparisonwithdeflare}. 
Note that CLAHE~\cite{Zuiderveld94} and BBPNet~\cite{singh2020single} are trained with the cross-entropy loss and triplet loss, we remove the IC loss in our FACENet for a fair comparison.
Although CLAHE~\cite{Zuiderveld94} and BBPNet~\cite{singh2020single} slightly improve the accuracy to some content, they reduce the flare corruption by suppressing the background noise and restoring the information according to randomly added flare, therefore leading to limited performance. 
By jointly predicting the flare masks and locally guiding the flare-immunized information to RGB and NI, FACENet significantly beats CLAHE~\cite{Zuiderveld94} and BBPNet~\cite{singh2020single} by a large margin.
%
%
It is worth noting that BPPNet~\cite{singh2020single} requires paired training data, while CLAHE~\cite{Zuiderveld94} and our method do not.
\begin{table}[!ht]
\setlength{\tabcolsep}{1.2mm}
\centering
\caption{Comparison with the de-flare methods to decrease the influence of intense extra light. 
Ex\_Tra indicates if extra training data is required (in \%).
}
\begin{tabular}{l|c|c|c|c|c}
\hline
 Method     & Ex\_Tra & mAP  & R-1 & R-5 & R-10 \\ \hline
Baseline    & - & 53.7                         & 60.3                         & 69.1                       & 74.4\\
 + CLAHE~\cite{Zuiderveld94} & - & 54.5 & 60.5  & 70.8  & 74.8   \\ 
 + BBPNet~\cite{singh2020single}& \ding{51} & 55.8 & 61.6  & 71.7  & 75.8   \\ 
\textbf{+  MFMP + FCE }   & -  & \bf60.5    &  \bf66.5     & \bf76.2   & \bf79.7 \\
\hline
\end{tabular}

\label{tab:comparisonwithdeflare}
\end{table}

\noindent
{\textbf{Plugging to different layers of the backbone.}}
To explore the best performance of our method in different layers, we plug in the proposed MFMP and FCE into four different layers respectively.
As shown in Table~\ref{tab:pluginlayer}, our method benefits from the deeper layer of ResNet50.
The main reason is the richer semantic information in the deep layer of ResNet50, the more accurate flare mask prediction, leading to better flare-aware cross-modal enhancement.

\begin{table}[ht]
\caption{Experimental results of plugging proposed MFMP and FCE into different layers of ResNet50 (in \%).}
\centering
\begin{tabular}{l|c|c|c|c}
\hline
Plugin Layer & mAP  & R-1 & R-5 & R-10 \\ \hline
Layer\_1     & 50.6    & 56.8     & 68.5      & 73.0       \\
Layer\_2     & 53.8 & 60.5   & 71.2   & 76.2    \\
Layer\_3     & 55.5 & 60.7   & 72.6   & 75.6    \\
Layer\_4      &60.5    & 66.5     &76.2   &79.7    \\ \hline
\end{tabular}

\label{tab:pluginlayer}
\end{table}

\subsection{Evaluation on IC Loss: $\mathcal{L}_{IC}$}
\noindent
\textbf{Comparing with the state-of-the-art loss functions.}
We further compare our IC loss with the state-of-the-art loss functions as shown in Table~\ref{tab:loss_comp}. 
%
%
%
3M loss~\cite{wang2022interact} enforces TI differently from the TI-guided RGB and NI features, which suppresses the effectiveness of MFMP and FCE modules.
CdC loss~\cite{zheng2022multi} enforces RGB and NI more similar to TI features, thus losing their original discriminative power.
HC loss~\cite{zhu2020hetero} is originally designed for cross-modal Re-ID by constraining the consistency of RGB and NI spectra.
Implying HC loss in our task explores the common effective features in RGB and NI, therefore outperforming 3M loss and CdC loss.
However, the performance is still behindhand, since HC loss aims to maximize the similarity of feature centers, which may excessively focus on the features that are enhanced by the TI spectrum.
By contrast, the proposed IC loss explores the semantic similarity according to prediction scores of RGB and NI spectra, thus leading to a significant performance improvement.

\begin{table}[H]
\caption{Comparison with existing modality relationship loss functions (in \%).}
\label{tab:loss_comp}
\begin{tabular}{l|c|c|c|c}
\hline
             & mAP & R-1 & R-5 & R-10                   \\ \hline
FACENet w/o $\mathcal{L}_{IC}$  & 60.5    &  66.5     & 76.2   & 79.7       \\ \hline
+ $\mathcal{L}_{3M}$~\cite{wang2022interact}    & 50.3   & 55.3   & 70.3   & 77.4                      \\ \hline
+ $\mathcal{L}_{HC}$~\cite{li2020multi}    &59.4     &64.9    &75.4    &79.5                       \\ \hline
+ $\mathcal{L}_{CdC}$~\cite{zheng2022multi}   &54.1     &58.2    & 70.1   & 74.7                      \\ \hline
+ $\mathcal{L}_{IC}$  & 62.1   &  67.3    & 76.7     & 80.1    \\ \hline
\end{tabular}
\end{table}

\noindent
\textbf{Plugging to state-of-the-art multi-modal Re-ID methods.}
To evaluate the effectiveness of our proposed IC loss, we plug it into different multi-modal Re-ID methods on WMBeID863 as shown in Table \ref{tab:iclossplugin}. 
It is obvious that plugging IC Loss into existing methods can effectively align the semantic information of RGB and NI, thus substantially boosting the performance.

\begin{table}[H]
\caption{Experimental results of plugging the proposed IC Loss $\mathcal{L}_{IC}$ into existing multi-modal Re-ID methods on WMVEID863 (in \%).}
\label{tab:iclossplugin}
\begin{tabular}{l|c|c|c|c}
\hline
       & mAP & R-1 & R-5 & R-10 \\ \hline
HAMNet~\cite{li2020multi}  & 45.6 & 48.5  & 63.1  & 68.8    \\ 
+ $\mathcal{L}_{IC}$  &48.8     &52.1        & 67.8 &  72.1   \\ \hline
PFNet~\cite{zheng2021robust}      & 50.1 & 55.9  & 68.7  & 75.1   \\ 
+ $\mathcal{L}_{IC}$ &54.0     &58.5    &72.1    & 76.9    \\ \hline
IEEE~\cite{wang2022interact}   & 45.9 & 48.6  & 64.3  & 67.9   \\ 
+ $\mathcal{L}_{IC}$ & 51.4    & 59.3   &  72.6  &  75.1   \\ \hline
CCNet~\cite{zheng2022multi}    &{50.3} &{52.7} &{69.6} &{75.1}   \\ 
+ $\mathcal{L}_{IC}$ &60.6&68.5&79.2&82.4   \\ \hline
\end{tabular}
\end{table}


\noindent

\subsection{Comparison with State-of-the-Art Methods}
\noindent
We perform a comprehensive evaluation on WMVeID863, as shown in Table~\ref{tab:comparisonwithsota}.
%
Most existing single-modal methods fail to utilize the complementary information of multi-modal, and thus cannot overcome the influence of intense flare.
It is worth noting that the transformer-based methods~\cite{he2021transreid, dosovitskiy2020image} significantly outperform CNN-based methods, the reason is that the patch-based transformers discover both global and local parts of an image, thus can better model the connection between the flare-corrupted region and flare-immunized region to suppress the influence of intense flare.
However, patch-based transformers are usually very computationally intensive.
%
%
%
Existing multi-modal methods aim to fuse complementary features, however, cannot recover the information corrupted by the intense flare.
By contrast, FACENet focuses on using flare-immunized information to guide the feature learning of flare-susceptible spectra, thus leading to the best overall feature representation.

\begin{table}[!h]
\setlength{\tabcolsep}{1.2mm}
\centering
\caption{Comparison with State-of-the-Art methods for multi-spectra Re-ID on WMVeID863 (in \%). The best three scores are marked in \textcolor{red}{red}, \textcolor{blue}{blue} and \textcolor{green}{green} (in \%). }
\begin{tabular}{l|c|c|c|c|c}
\hline
                              & Method         & mAP & R-1 & R-5 &  R-10 \\ \hline
\multirow{8}{*}{Single-Modal}  & DenseNet~\cite{huang2017densely}     &  42.9   &  47.9  & 61.9   &  68.7   \\            
                              & ShuffleNet~\cite{zhang2018shufflenet}       &   34.2  & 37.2   &  52.3  & 58.9    \\ 
                              & MLFN~ \cite{chang2018multi}      &43.7     & 47.0   &  61.7  &   69.8  \\ 
                              & HACNN~\cite{li2018harmonious}       &46.9 &48.9     & 66.9   &   73.8     \\ 
                              & StrongBaseline~\cite{luo2019bag}  &51.1 &55.7 &69.8 &74.7    \\
                              & OSNet~\cite{zhou2019omni}       &42.9     &46.8    &61.9    & 69.4    \\ 
                              & ViT~ \cite{dosovitskiy2020image}& \textcolor{green}{54.7} & \textcolor{green}{61.2} & \textcolor{green}{72.8} & \textcolor{green}{78.1}\\
                              & TransReID~\cite{he2021transreid}       &\textcolor{blue}{59.4} &\textcolor{blue}{66.5} & \textcolor{blue}{76.3} & \textcolor{blue}{80.1}    \\
                              & AGW~\cite{ye2021deep}            & 30.3    & 35.3   & 43.3   & 46.5    \\
                              & PFD~\cite{wang2022pose} &50.2 &55.3 &69.8 &75.3\\

                              \hline
\multirow{5}{*}{Multi-Modal}  & HAMNet~\cite{li2020multi}         & 45.6 & 48.5  & 63.1  & 68.8     \\
                              & PFNet~\cite{zheng2021robust}          & 50.1 & 55.9  & 68.7  & 75.1   \\
                              & IEEE~\cite{wang2022interact}           & 45.9 & 48.6  & 64.3  & 67.9     \\
                              & CCNet~\cite{zheng2022multi}          &{50.3} &{52.7} &{69.6} &{75.1} \\
                              & {\bf FACENet}           & \textcolor{red}{62.1}   & \textcolor{red} {67.3}    & \textcolor{red}{76.7}     & \textcolor{red}{80.1}     \\ \hline
\end{tabular}

\label{tab:comparisonwithsota}
\end{table}

\subsection{Evaluation on Different Baselines}
Table~\ref{tab:comparisonwithsota} shows that the
transformer-based methods ViT~\cite{dosovitskiy2020image} and TransReID~\cite{he2021transreid} present advanced performance comparing to the other state-of-the-art methods.
To further validate the effectiveness of our method, we integrate the key components MFMP, FCE, and IC loss from our FACENet into ViT~\cite{dosovitskiy2020image} and TranReID~\cite{he2021transreid}.
As shown in Table~\ref{tab:diffbackbones}, all the metrics improve after integrating our method into the different baselines, which indicated the generality of our method.

\begin{table}[H]
\caption{Evaluation of the proposed method on different backbones, where Ours = MFMP + FCE + $\mathcal{L}_{IC}$.}
\label{tab:diffbackbones}
\begin{tabular}{l|c|c|c|c}
\hline
            & mAP  & R-1  & R-5  & R-10 \\ \hline
ResNet50~\cite{he2016deep}      & 53.7 & 60.3 & 69.1 & 74.4 \\ \hline
+ Ours & 62.1 & 67.3 & 76.7 & 80.1 \\ \hline
ViT~\cite{dosovitskiy2020image}         & 54.7 & 61.2 & 72.8 & 78.1 \\ \hline
+ Ours    & 58.4 & 62.6 & 74.6 & 79.9 \\ \hline
TransReID~\cite{he2021transreid}         & 59.4 &66.5 &{76.3} & {80.1} \\ \hline
+ Ours    & 62.9 & 70.5 & 78.6 & 82.6 \\ \hline
\end{tabular}
\end{table}

\section{Conclusion}
In this paper, we first investigate the intense flare challenge in vehicle Re-ID, which results in a significant drop in image quality in certain spectra.
To address the problem, we propose a novel Flare-Aware Cross-modal Enhancement Network (FACENet), which accurately localizes the flare-corrupted area in a self-supervised manner and enhances the flare-susceptible spectra with the guidance of the flare-immuned spectrum in a conditional learning manner. 
In addition, we propose a simple yet effective inter-modality consistency loss to align the semantic information of the flare-susceptible spectra, which further utilizes the useful information under the intense flare.
Moreover, we contribute a new dataset WMVeID863 with more realistic challenges such as intense flare and motion blur for multi-spectral vehicle Re-ID and related communities.
Comprehensive experimental results on the proposed WMVeID863 demonstrate the promising performance of the proposed method FACENet, especially while handling intense flares.
%

\printbibliography

\end{document}